\documentclass[letterpaper, 10 pt, conference]{ieeeconf}  % Comment this line out if you need a4paper

\IEEEoverridecommandlockouts                              % This command is only needed if 
\usepackage{amsmath} % assumes amsmath package installed
\usepackage{amssymb}  % assumes amsmath package installed
\usepackage{graphicx}
\usepackage[font=footnotesize]{caption}
\usepackage[font=footnotesize]{subcaption}
\usepackage[colorinlistoftodos,prependcaption,textsize=tiny]{todonotes}

\usepackage{url}
\usepackage{flushend}

\title{\LARGE \bf
CUREE: A Curious Underwater Robot for Ecosystem Exploration
}

\author{Yogesh Girdhar$^{1}$, Nathan McGuire$^{1}$, Levi Cai$^{2}$, Stewart Jamieson$^{2}$, Seth McCammon$^{1}$, Brian Claus$^{1}$,\\  John E. San Soucie$^{2}$, Jessica E. Todd$^{2}$, and T. Aran Mooney$^{1}$
\thanks{*This work was supported in part by NSF Grants 1734400 and 2133029, the Investment in Science Fund at WHOI, the Postdoctoral Scholar Program at WHOI, the NDSEG Fellowship Program, and by grants from NVIDIA. It utilized an NVIDIA RTX A6000.}% <-this % stops a space
\thanks{$^{1}$Y. Girdhar, N. McGuire, S. McCammon, B. Claus, and T. A. Mooney are with the Woods Hole Oceanographic Institution (WHOI)
        {\tt\small \{yogi, nmcguire, smccammon, bclaus, amooney\}@whoi.edu}}%
\thanks{$^{2}$L. Cai, S. Jamieson, J. E. San Soucie, and J. E. Todd are with the MIT-WHOI Joint Program in Oceanography/Applied Ocean Science and Engineering
        {\tt\small \{lcai, sjamieson, jsansoucie, jtodd\}@whoi.edu}}%
}

\begin{document}

\maketitle
\thispagestyle{empty}
\pagestyle{empty}

%%%%%%%%%%%%%%%%%%%%%%%%%%%%%%%%%%%%%%%%%%%%%%%%%%%%%%%%%%%%%%%%%%%%%%%%%%%%%%%%
\begin{abstract}
%%%%%%%%%%%%%%%%%%%%%%%%%%%%%%%%%%%%%%%%%%%%%%%%%%%%%%%%%%%%%%%%%%%%%%%%%%%%%%%%

The current approach to exploring and monitoring complex underwater ecosystems, such as coral reefs, is to conduct surveys using diver-held or static cameras, or deploying sensor buoys. These approaches often fail to capture the full variation and complexity of interactions between different reef organisms and their habitat. The CUREE platform presented in this paper provides a unique set of capabilities in the form of robot behaviors and perception algorithms to enable scientists to explore different aspects of an ecosystem. Examples of these capabilities include low-altitude visual surveys, soundscape surveys, habitat characterization, and animal following. We demonstrate these capabilities by describing two field deployments on coral reefs in the US Virgin Islands. In the first deployment, we show that CUREE can identify the preferred habitat type of snapping shrimp in a reef through a combination of a visual survey, habitat characterization, and a soundscape survey. In the second deployment, we demonstrate CUREE's ability to follow arbitrary animals by separately following a barracuda and stingray for several minutes each in midwater and benthic environments, respectively. 

\end{abstract}

%%%%%%%%%%%%%%%%%%%%%%%%%%%%%%%%%%%%%%%%%%%%%%%%%%%%%%%%%%%%%%%%%%%%%%%%%%%%%%%%
\section{INTRODUCTION}
\label{sec:intro}
%%%%%%%%%%%%%%%%%%%%%%%%%%%%%%%%%%%%%%%%%%%%%%%%%%%%%%%%%%%%%%%%%%%%%%%%%%%%%%%%

Robots have successfully been deployed for extensive adaptive monitoring of underwater ecological phenomena, such as toxic cyanobacteria \cite{Hitz2017}, phytoplankton \cite{Fossum2019, Das2015a}, tracking animals in the mid-water column \cite{benoit2018equipping, Yoerger2021}, and benthic surveys of coral reef-like environments \cite{Manderson2020} \cite{Armstrong2008, Singh2007}.  However, their use has either been limited to mapping tasks over the seafloor, or informative path planning over simple scalar observations using mission specific sensors in the mid water column. Coral reef ecosystems consist of complex seafloor geometry, a large number of species, and the manifold of interactions between organisms and their habitats.  To monitor reefs, scientists still primarily rely on divers to conduct survey missions.  Autonomous Underwater Vehicles (AUVs) offer an unparalleled ability to measure phenomena of scientific interest distributed across space and time. However, to fully observe these ecosystems, robots must develop an understanding of different components of the ecosystem and use these components to perform more targeted hypothesis-driven data collection. This paper presents the Curious Underwater Robot for Ecosystem Exploration (CUREE), a robotic system that provides ecologically relevant behaviors and perception subsystems that can be used as building blocks to rapidly construct missions to build understanding of different aspects of the ecosystem. 

\begin{figure}
    \centering
    \includegraphics[width=\linewidth]{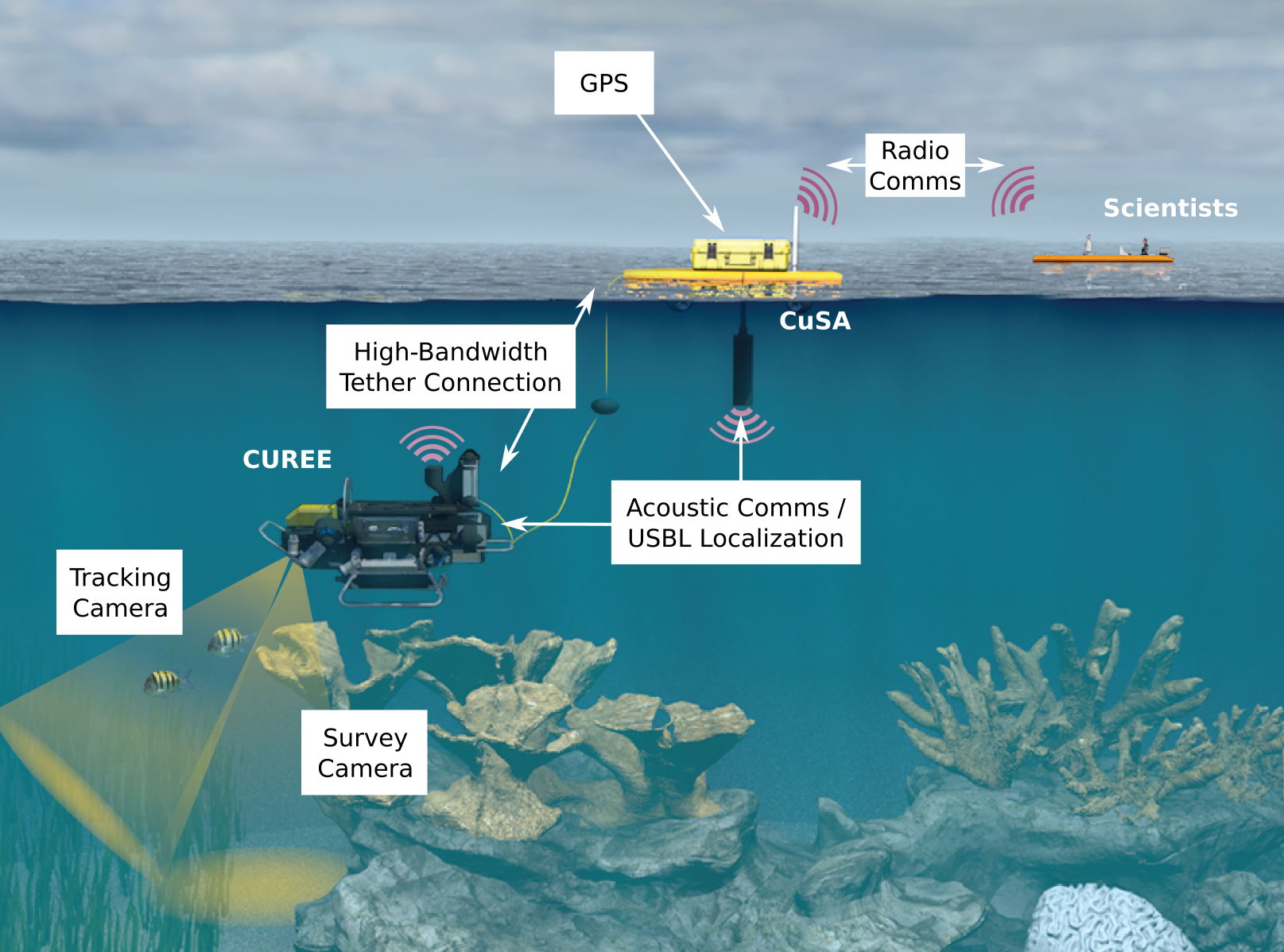}
    \caption{Illustration of CUREE conducting a reef survey.  CUREE uses vision and passive acoustics to collect information about its environment.  CuSA assists CUREE by providing a high-bandwidth communications link to scientists and improved localization of CUREE with GPS and USBL.}
    \label{fig:system_overview}
\end{figure}

Images are perhaps the best approach for capturing species-level information about organisms without destructive sampling \cite{Richards2019}. However, existing approaches to computer vision underwater are still very noisy. Even in the best case scenario, vision can only be used to make observations within a small range (\textless 5~m) of the robot location. Passive acoustic observations made by hydrophones enable the robot to make observations over much longer ranges.  However, these acoustic observations do not contain as much species level information, since not every animal makes measurable sounds.  Recent efforts have shown the use of sound source localization (SSL) \cite{Rascon2017} techniques by robots for the purpose of target acquisition and tracking.  We hypothesize that integrating visual and acoustic sensors would be an ideal combination for a robot tasked with monitoring a complex ecosystem. Such a robot would primarily use images for species level observation, identify different habitat types, and for collision avoidance close to the seafloor, and use passive acoustics observation to approximate biodiversity and animal presence over longer spatial scales. 

The novel physical design of the robot was guided by identifying requirements for transportability and deployment to remote parts of the world, heat dissipation and computation associated with running state-of-the-art perception and planning algorithms, need for in-water behaviors such as drifting, and enabling continuous rapid research and development efforts through a modular design. 

CUREE can be optionally accompanied by CUREE's Surface Assistant (CuSA), a surface vehicle which autonomously follows CUREE while acting as a radio communication relay and providing CUREE with its global position information (Fig.~\ref{fig:system_overview}). 

%Many recent robotic platforms, such as the BlueRobotics BlueROV2, OpenROV Trident, Bluefin SandShark UUV, RSE Guardian LF1 Mark 3, Aqua Amphibious robot, or QUT Rangerbot, have been built to provide easily deployable, vision-based alternatives to aid in scientific underwater tasks. Many of these platforms are closed-source, prohibitively expensive, or were developed for specific tasks. Only the BlueROV2 is both open-source and relatively affordable. However, the BlueROV2 can be difficult to transport and modify for specific tasks and payloads.

%%%%%%%%%%%%%%%%%%%%%%%%%%%%%%%%%%%%%%%%%%%%%%%%%%%%%%%%%%%%%%%%%%%%%%%%%%%
\section{Monitoring ecosystems: key capabilities}
\label{sec:key_capabilities}
%%%%%%%%%%%%%%%%%%%%%%%%%%%%%%%%%%%%%%%%%%%%%%%%%%%%%%%%%%%%%%%%%%%%%%%%%%%

The biological components of a coral reef ecosystem are made up of stationary plants, corals, and algae attached to the seafloor, as well as fish, turtles, other mobile animals, and microscopic organisms which move about in the water column. The geophysical component is defined in terms of seafloor substrate types, depth, and the physical and chemical properties of the water. CUREE aims to perform missions that target observations of different components of the ecosystem with its cameras and hydrophones. CUREE's core capabilities are summarized in the subsections below. 

% In this work we assume a simplified model of the ecosystem consisting of organisms that are visible with the robot's camera systems or can be heard with the its hydrophones.  We approximate the physical environment to visually distinguishable substrate types such as sand, rubble, or hard corals. 

\subsection{Low Altitude Benthic Surveys}

Conducting benthic surveys is perhaps the most common mode of operation for most AUVs. However, unlike standard bathymetric surveys with sonar devices, such as Doppler Velocity Loggers (DVLs) or echosounders, which can be performed while keeping the vehicle at high altitudes, the visual benthic surveys used to assess coral and fish biodiversity require AUVs to operate close to the sea floor to obtain high-resolution imagery. Acoustic sensors work well when there is a hard, mostly flat bottom. However, they often fail in the presence of soft or narrow objects such as soft corals and plants that are commonly encountered by AUVs operating at \textless1.5m altitude in coral reefs. 

To enable low-altitude observations we rely on the vehicle's DVL to provide a robust estimate of the distance between CUREE and the seafloor, up to ~1.5m.  By fusing the altitude estimate with local position information from CUREE's DVL and IMU, and with global positioning from its USBL, CUREE can localize itself in complex reef-like environments. 

\begin{figure}
    \centering
    \includegraphics[width=\linewidth]%{figs/tektite_reef.jpg}
    {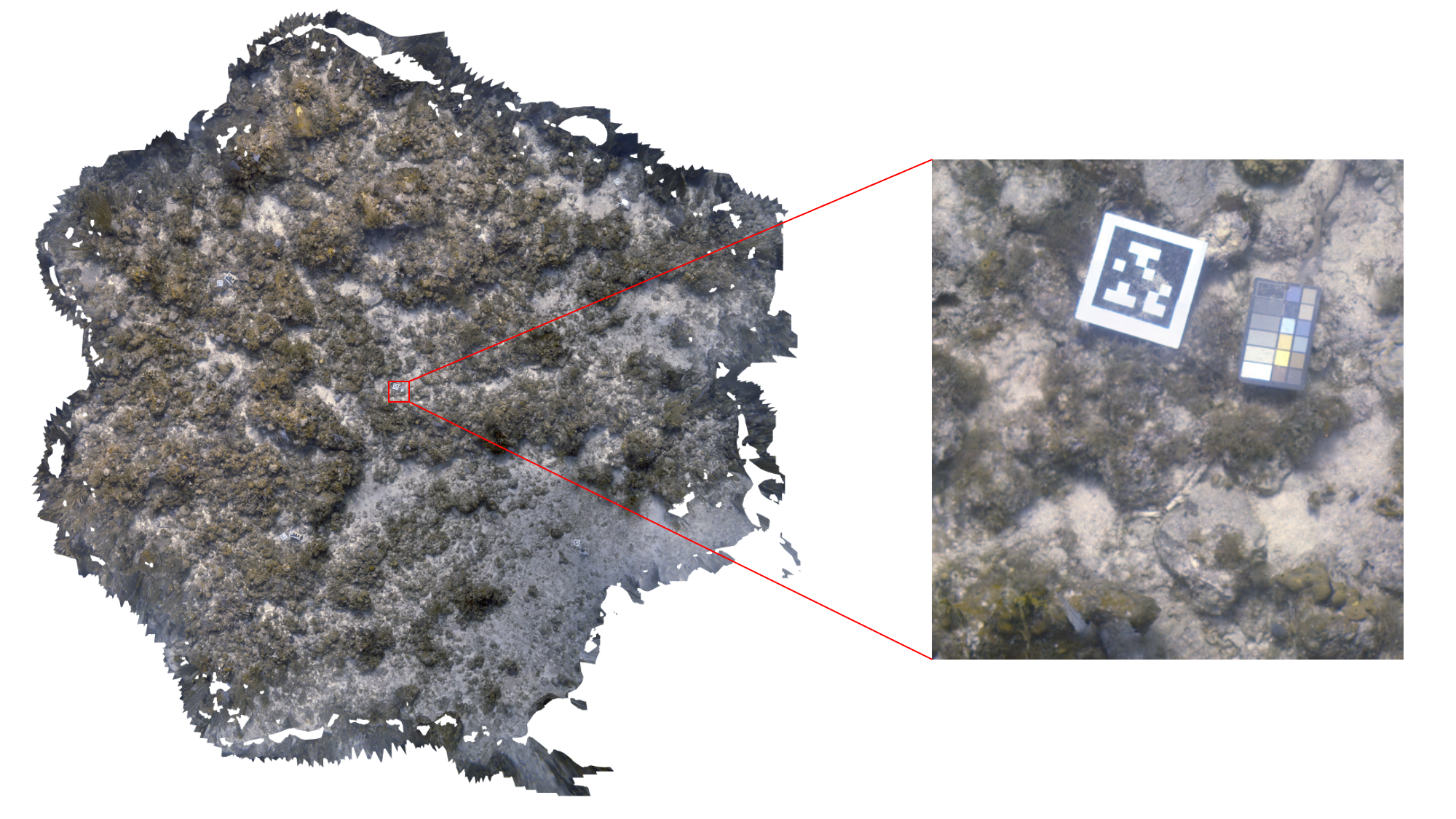}
    \caption{3D reconstruction of Booby Rock Reef, St. John, USVI, produced from a CUREE visual survey. The AprilTag~\cite{Olson2011} shown in the cutout is 20cm $\times$ 20cm, while the complete survey is approximately 20m $\times$ 20m.}
    %\caption{3D Reconstruction of Tektite Reef, St. John, USVI using data collected by CUREE.  Total mission duration was 19 minutes 40 seconds. The reconstructed area is approximately 12m x 12m}
    \label{fig:tektite_benthic_survey}
\end{figure}

Similarly, CUREE can localize the high-resolution imagery as well. We can then process this imagery offline to produce 3D representations of the reef \cite{ferrari2017}, which is represented by the distribution $P(\mathrm{img} | x)$.
An example of such a 3D reef survey is shown in Fig.~\ref{fig:tektite_benthic_survey}. 

However, we anticipate future experiments where CUREE would need to operate even closer to the seafloor (e.g. to take a water sample near a specific coral head) or near narrow objects, such as pillar coral.  Doing so in soft or narrow terrains may make the DVL estimate unreliable.  As an alternative CUREE can produce an altitude estimate created from stereo-vision to provide a high-resolution alternative to the DVL estimate.

\subsection{Unsupervised Substrate Characterization}
Ecologists often characterize reef habitats in terms of different substrate types such as coral, rubble, algae, and sand, and how they are distributed in space. The task of discovering and mapping different habitat types can be understood as learning the factorization of the spatial distribution of the visual seafloor observations $P(\mathrm{img} | x)$ in terms of an appearance model of each habitat distribution $P(\mathrm{img} | \mathrm{habitat})$ and the spatial distribution of different habitats $P(\mathrm{habitat} | x)$ : 

\begin{multline*}
P(\mathrm{img} | x) = \sum_{\mathrm{habitats}} P(\mathrm{img} | \mathrm{habitat}) P(\mathrm{habitat} | x). 
\end{multline*}

Giving AUVs the ability to discover habitat types and provide habitat labels to locations in real-time can enable observations of organisms to be understood in the context of their habitat, identifying anomalous migrations.  

However, factors such as bad visibility, presence of different underlying species or corals or algae, overlapping categories, altitude, and the varying health condition of the reef can make it hard to categorically label the observed seafloor using a supervised neural network classifier. Moreover, moving beyond coral reefs, there exist many different types of substrate types for which there is insufficient prior data to train a convolution neural net (CNN). 

Instead of using a pre-trained supervised classifier to detect specific habitat types, CUREE uses an unsupervised approach to automatically discover different habitat types in realtime, and use them to produce a continuously updating habitat map $P(\mathrm{habitat} | x)$. The approach uses a Hierarchical Dirichlet Process (HDP) based topic model \cite{Girdhar2019}, which does inference in realtime using a Gibbs sampler. The unsupervised learning and realtime inference properties of this approach enables CUREE to be deployed in arbitrary environments with no requirements on prior knowledge.

\subsection{Soundscape Surveys}
\begin{figure}
    \centering
    \includegraphics[width=.9\linewidth]{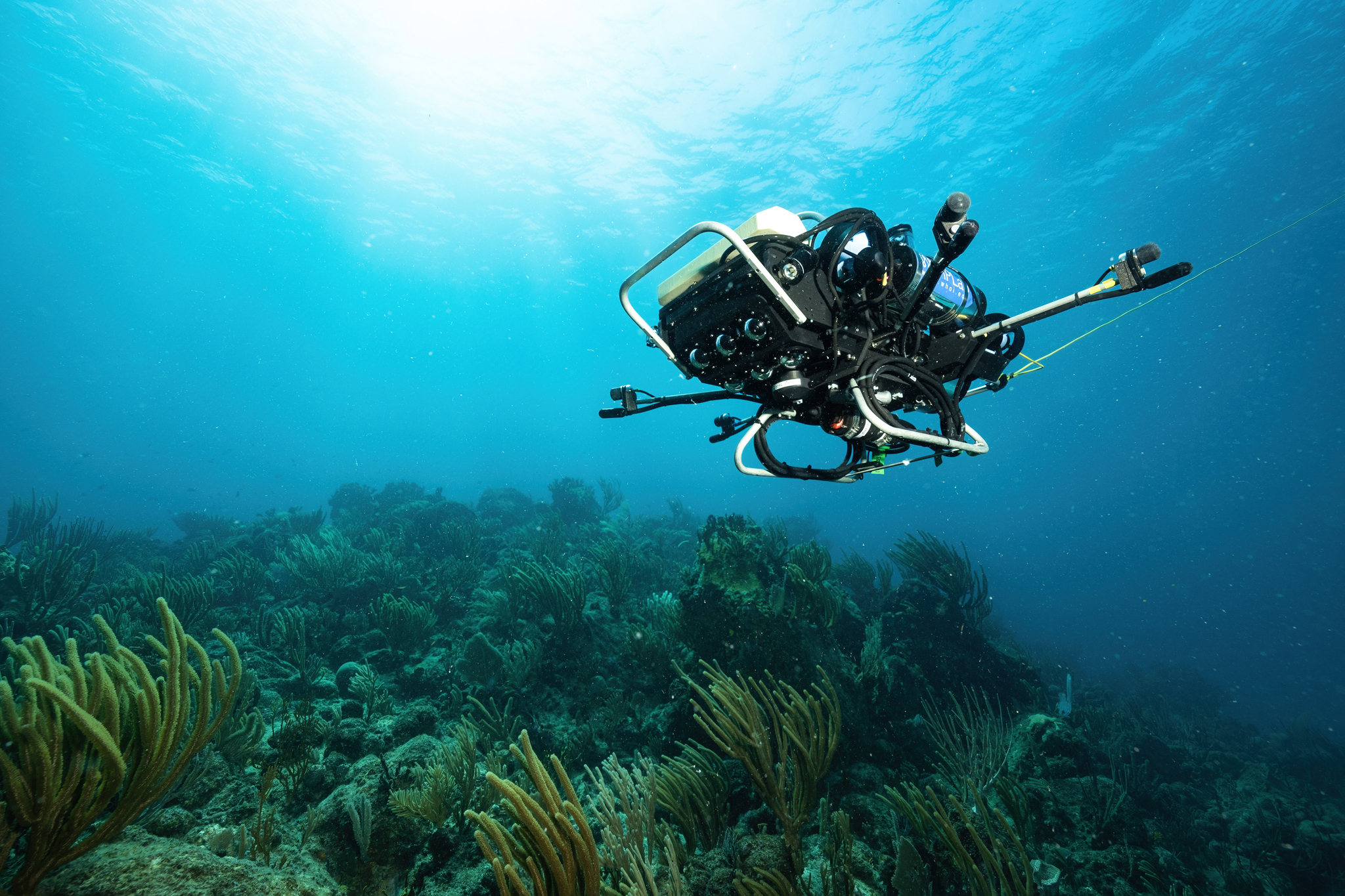}
    \caption{CUREE's primary sensing capabilities come from forward and downward-looking stereo cameras and a four-channel hydrophone array (Photo credit: Austin Greene).}
    \label{fig:warpauv}
\end{figure}

Marine soundscapes are composed of biological, geological, physical, and anthropogenic sound sources. While visual observations can provide dense and specific information about the robot's immediate surroundings, they suffer from several drawbacks.  Underwater cameras have a limited range and can only provide information about objects within the cameras field of view.  In contrast, hydrophones provide an omnidirectional sensor with much greater range.  Multiple hydrophones in an array enable Sound Source Localization (SSL) by comparing the time difference of arrival of sound at the different array elements, or through more complex signal analysis techniques, including beamforming \cite{urick1975principles}.  CUREE's hydrophone payload consists of four synchronized hydrophones positioned away from the robot's body at the end of aluminum arms as can be seen in Fig.~\ref{fig:warpauv}.  

\begin{figure}
    \centering
    \includegraphics[width=.8\linewidth]{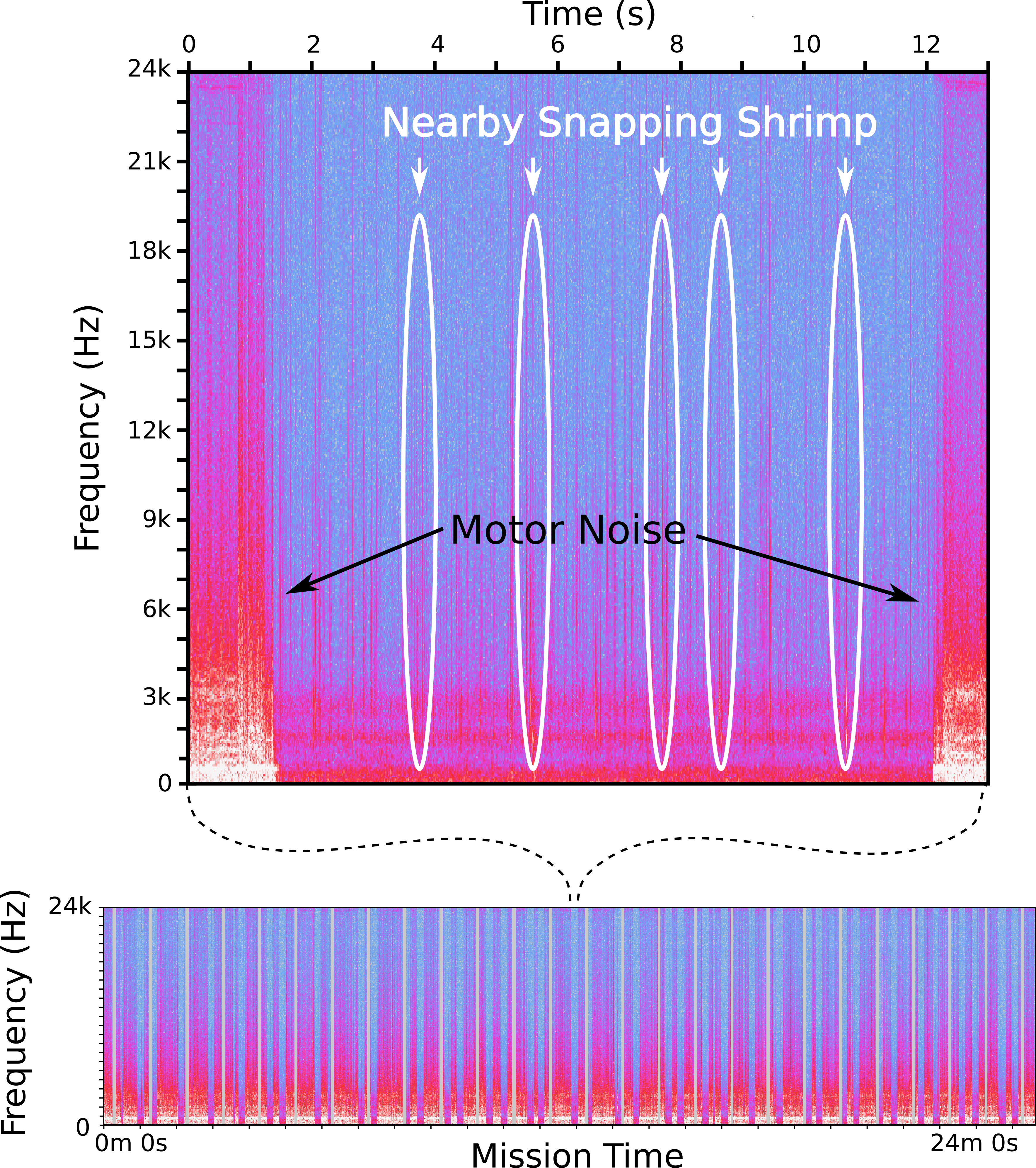}
    \caption{Sample Acoustic Survey conducted at Joel's Shoal Reef, US Virgin Islands. During drifting periods, there is no noise from the thrusters, allowing CUREE to detect sounds from marine animals, such as snapping shrimp.}
    \label{fig:acoustic_mission}
\end{figure}

Soundscapes observed over a coral reef show variability in both space and time due to diurnal animal activity, tides, and storms \cite{lillis2018snapping}. The observed distribution of acoustic features (spectrogram) can be represented by the distribution

\begin{equation*}
    P(\mathrm{sound} | x, t).
\end{equation*}

Previous attempts to capture the spatial variation of soundscapes have been through the deployment of fixed hydrophone arrays \cite{freeman2015cross}, drifting buoys equipped with hydrophones that passively drift across a region of interest \cite{lillis2018drifting}, or hydrophone-equipped gliders \cite{baumgartner2008diel, dassatti2011board, wall2012shelf}.  However, these approaches are limited in their effectiveness in capturing spatial variation.  Static arrays can only observe in one location and can only capture the direction of arrival of a sound, not the exact source location.  Since they can move through an environment, drifters and gliders can better capture the spatial variation of soundscapes. However, drifters' ability to cover an environment is dictated entirely by the ocean currents, and gliders cannot operate in the shallow waters of many coral reefs.  Due to its size and maneuverability, CUREE is capable of operating in these shallow, highly dynamic environments. However CUREE's thruster noise saturates its hydrophone's dynamic range making it difficult to collect useful acoustic data while moving. CUREE addresses this by interleaving periodic drifting behavior, free from thruster noise, during a regular benthic survey (Fig.~\ref{fig:acoustic_mission}). As a result, CUREE can make soundscape observations at arbitrary locations along a mission, enabling adaptive soundscape-guided missions.

%Fig.\ref{fig:soundscape_habitat} shows an example a combined visuo-acoustic survey.   %While CUREE is drifting, it has no actuation, and is instead moved around by the currents and its own buoyancy.  We found that limiting the drifting periods' length to ten seconds prevented CUREE from drifting too far from its intended position.  

\subsection{Animal Tracking}

CUREE provides a novel capability to follow arbitrary animals in the wild. Unlike tracking animals in the mid-water column, where traditional computer vision techniques like blob detection are sufficient to lock on to slow-moving targets (e.g. jellyfish or larvaceans) \cite{yoerger2021hybrid}, tracking animals on a coral reef requires the ability to separate fast-moving fish from complex backgrounds. CUREE addresses this problem through the use of a semi-supervised visual tracker \cite{Cai2022a}\cite{Hu2022}.
Given a single bounding box annotation of an animal in the robot's view, CUREE visually locks on and follows, enabling it to observe the spatiotemporal distribution of the animal 

\begin{equation*}
   P(x,t | \mathrm{animal})
\end{equation*}
over long periods of time.

These types of observations (long continuous tracks of organisms) are invaluable to marine biologists, since they provide a unique perspective on the animal's behavior that may not be replicable in a laboratory setting.  Stationary camera traps will fail to capture this distribution, since they can only observe $P(\text{species} | x)$. The primary other way to obtain these observations is to use human divers, an approach which scales poorly.  This visual tracker provides the robot with the relative position of the animal, which then is used by the tracking controller to move the robot to keep the target animal in the center of the robot's view.  

CUREE's animal tracking behavior, when coupled with other instances of the same or other behaviors, enables us to answer many interesting ecologically relevant questions. For example, correlating the spatiotemporal distribution of the animal with the habitat distribution gives us information about the animal's preferred habitats        
\begin{equation*}
   P(\mathrm{habitat} | \mathrm{animal}) = \sum_x P(\mathrm{habitat} | x) P(x | \mathrm{animal});
\end{equation*}
and correlating with observations of other animal tracks, assuming independent tracks of individual animals in the same region, gives us information about inter-species interactions
\begin{equation*}
   P(\mathrm{animal_1}, \mathrm{animal_2}) \approx \sum_{x} P(\mathrm{animal_1}|x) P(\mathrm{animal_2}|x).
\end{equation*}

%On CUREE, this capability is delivered by integrating a Nvidia Jetson, a powerful GPU-enabled computer into the modular head of the vehicle.  With access to a GPU for real-time processing of images, CUREE can utilize state-of-the-art Artificial Neural Networks, such as SiamMask \cite{Hu2022} and YOLOv5 \cite{glenn_jocher_2022_7002879} at up to 15 fps.  

%\subsection{Animal biodiversity characterization}
%might bias their behavior in presence of a robot \cite{Campbell2021}, cannot be done using straightforward application of waypoint survey based approaches. 

%%%%%%%%%%%%%%%%%%%%%%%%%%%%%%%%%%%%%%%%%%%%%%%%%%%%%%%%%%%%%%%%%%%%%%%%%%%%%%%%
\section{Field Demonstrations on Coral Reefs}
\label{sec:field_demos}
%%%%%%%%%%%%%%%%%%%%%%%%%%%%%%%%%%%%%%%%%%%%%%%%%%%%%%%%%%%%%%%%%%%%%%%%%%%%%%%%

CUREE has been field-tested in dozens of deployments on reefs around St. John in the U.S. Virgin Islands.  In this section, we describe two deployments that demonstrate CUREE's effectiveness in evaluating ecosystem-related hypotheses and monitoring dynamic phenomena. The first of these is an Audio-Visual survey of a coral reef conducted at Joel's Shoal.  This survey combines the unsupervised substrate characterization and soundscape survey to learn the preferred habitat of snapping shrimp. The second deployment was conducted at Tektite Reef, where CUREE demonstrated autonomous animal tracking by following a barracuda and a stingray, for several minutes, after CUREE was provided with an initial bounding box for each.

\subsection{Audio-Visual Benthic Survey}
\begin{figure*}
    \centering
    \includegraphics[width=.9\linewidth]{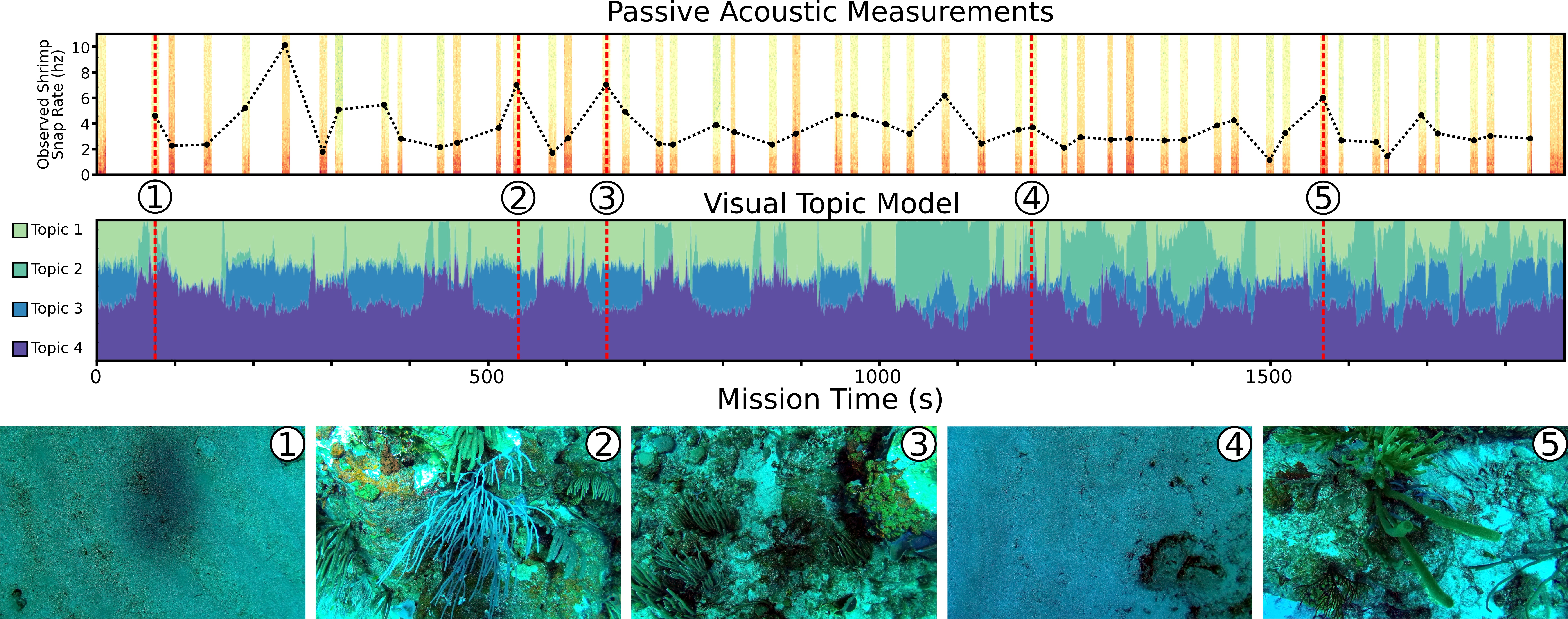}
    \caption{CUREE is capable of conducting simultaneous visual and acoustic surveys over complex seafloor environments such as a coral reefs. The approach mixes low-altitude terrain and waypoint following behavior with drifting periods to capture soundscapes. In this survey of Joel's Shoal in the USVI, CUREE drifts for 10 seconds at every waypoint to enable soundscape observation at that location. Top: Spectrogram of audio observations at each waypoint and average observed snapping shrimp snap rate during the drifting window. Middle: Automatically computed visual topic labels representing different habitat types. Bottom: Examples of imagery captured by the cameras at different points of time. We see that images 2, 3, and 5 have similar topic distribution and correspond to a coral-covered habitats, whereas images 1 and 4 have similar topic distributions (but distinct from 2,3,5) and show sandy habitats. We see that Topic 3 (blue), which indicates the presence of dense corals in the images, correlates positively with high rates of snapping shrimp snaps.}
    \label{fig:soundscape_habitat}
\end{figure*}

In this experiment, we deployed CUREE at Joel's Shoal reef in USVI, and conducted an audiovisual survey with 9370 images, interleaved with 50 drifting soundscape observations. The goal of this deployment was to collect a dataset to test whether it is possible to predict the presence of snapping shrimp using purely visual information. Snapping shrimp are known for their continuous high-frequency snaps, and their presence is mainly observed acoustically as they hide in rock crevices and burrows \cite{DUFFY2010421}. To simplify, we ignore acoustic multipath effects, though these can be considered in future experiments with the multi-hydrophone configuration.

\begin{figure}
    \centering
    \includegraphics[width=.7\linewidth]{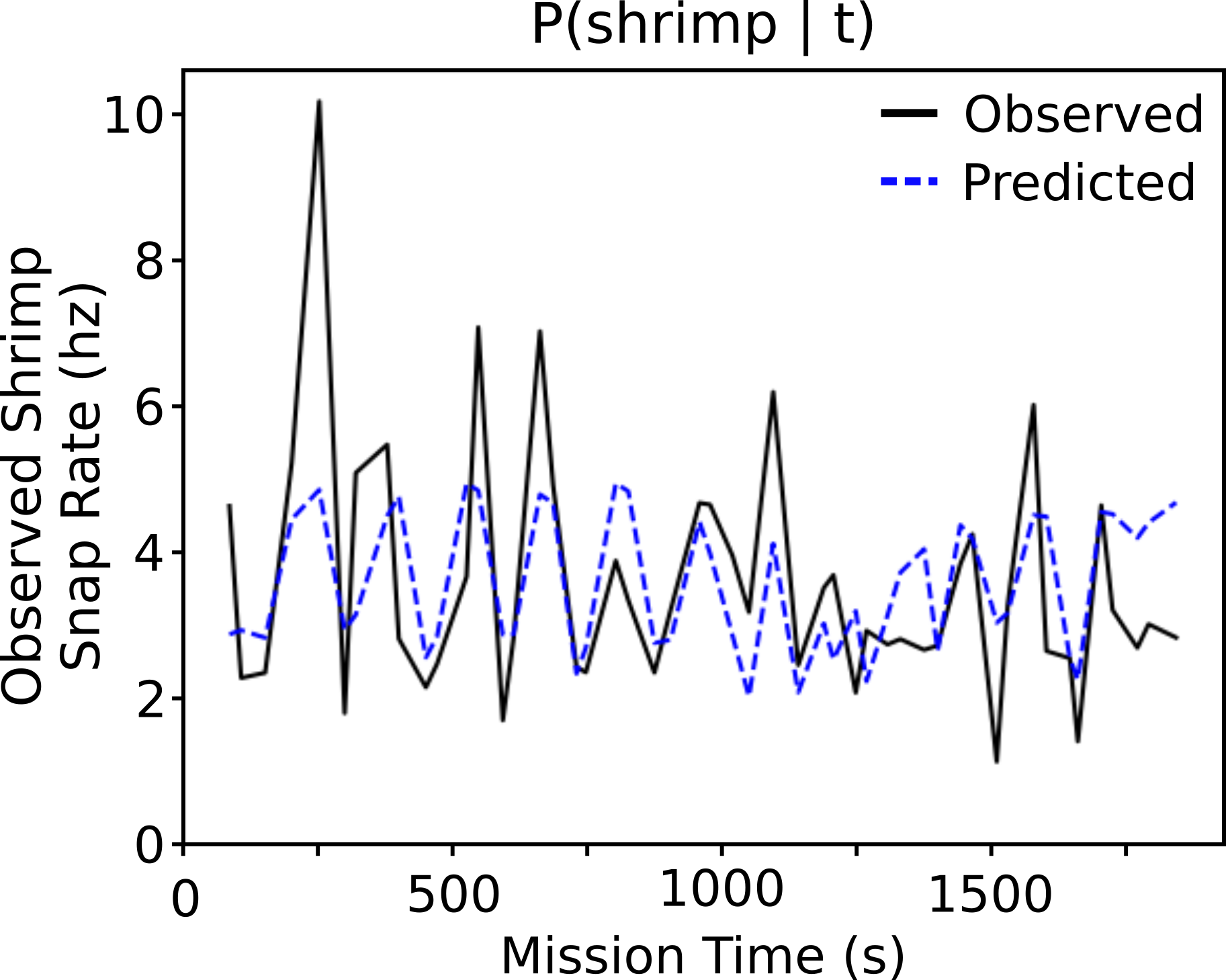}
    \caption{Plot shows normalized snapping shrimp snap rate as computed for soundscape observations (black), and the least squared linear fit of visual topic distribution to the normalized snap rate observations (blue dashed). We see that there is a strong correlation between the observed shrimp snaps and the predicted ones.}
    \label{fig:soundscape_habitat_corr}
\end{figure}

Given the observed soundscape, we first built an acoustic snapping shrimp snap detector using a method similar to \cite{radford2008temporal}, where we identify transient spikes in acoustic activity within the shrimp band (2~kHz to 24~kHz) above a threshold of 0.1$\sigma$ over the mean. The resultant counts of shrimp snaps form the distribution $P(\mathrm{shrimp} | \mathrm{sound})$.  Given this shrimp detector and the soundscape survey, we then estimate the snap distribution over time as
\begin{equation*}
    P(\mathrm{shrimp}| t) = \sum_{\mathrm{sound}} P(\mathrm{shrimp}| \mathrm{sound}) P(\mathrm{sound} | t).
\end{equation*}
This distribution is visualized as the black dashed line in the upper plot of Fig.~\ref{fig:soundscape_habitat} and the black line in Fig.~\ref{fig:soundscape_habitat_corr}.

Alternatively, we assume a generative model for snap distribution which only assumes knowledge of a timeseries of visual observations:\
\begin{multline*}
    P(\mathrm{shrimp}| \mathrm{img}, t) \\ = \sum_{\mathrm{habitat}} P(\mathrm{shrimp}| \mathrm{habitat}) P(\mathrm{habitat} | \mathrm{img}, t),
\end{multline*}
where $P(\mathrm{habitat} | img, t)$ is computed offline using HDP-ROST \cite{Girdhar2019}.  $P(\mathrm{habitat} | \mathrm{img}, t)$ can be seen in the middle plot in Fig.~\ref{fig:soundscape_habitat}, where each color in the time series corresponds to the relative prevalence of that habitat type in the imagery collected at time $t$.  From a qualitative examination of the images associated with each topic, we were able to determine that Topic 3, shown in blue, corresponded with observations of coral-covered regions. $P(\mathrm{shrimp}| \mathrm{habitat})$ is modeled by a linear least squares regression mapping habitat distribution to $P(\mathrm{shrimp}| \mathrm{sound})$.  The temporal snap distribution $P(\mathrm{shrimp}| t)$ is plotted as the blue dashed line in Fig.~\ref{fig:soundscape_habitat_corr}.

We find that the snap density as predicted by only visual observations correlates positively with the snap density computed using the soundscape data. The only topic for which there was a positive coefficient in the linear regression is Topic 3, the topic associated with rocky coral habitat.  While more experiments are necessary, this supports the hypothesis that snapping shrimp prefer the dense corals around Joel's Shoal as their habitat, where there are many more rock crevices and tiny spaces to hide in.

\subsection{Tracking Organisms}
\begin{figure}
    \centering
    \includegraphics[width=.9\linewidth]{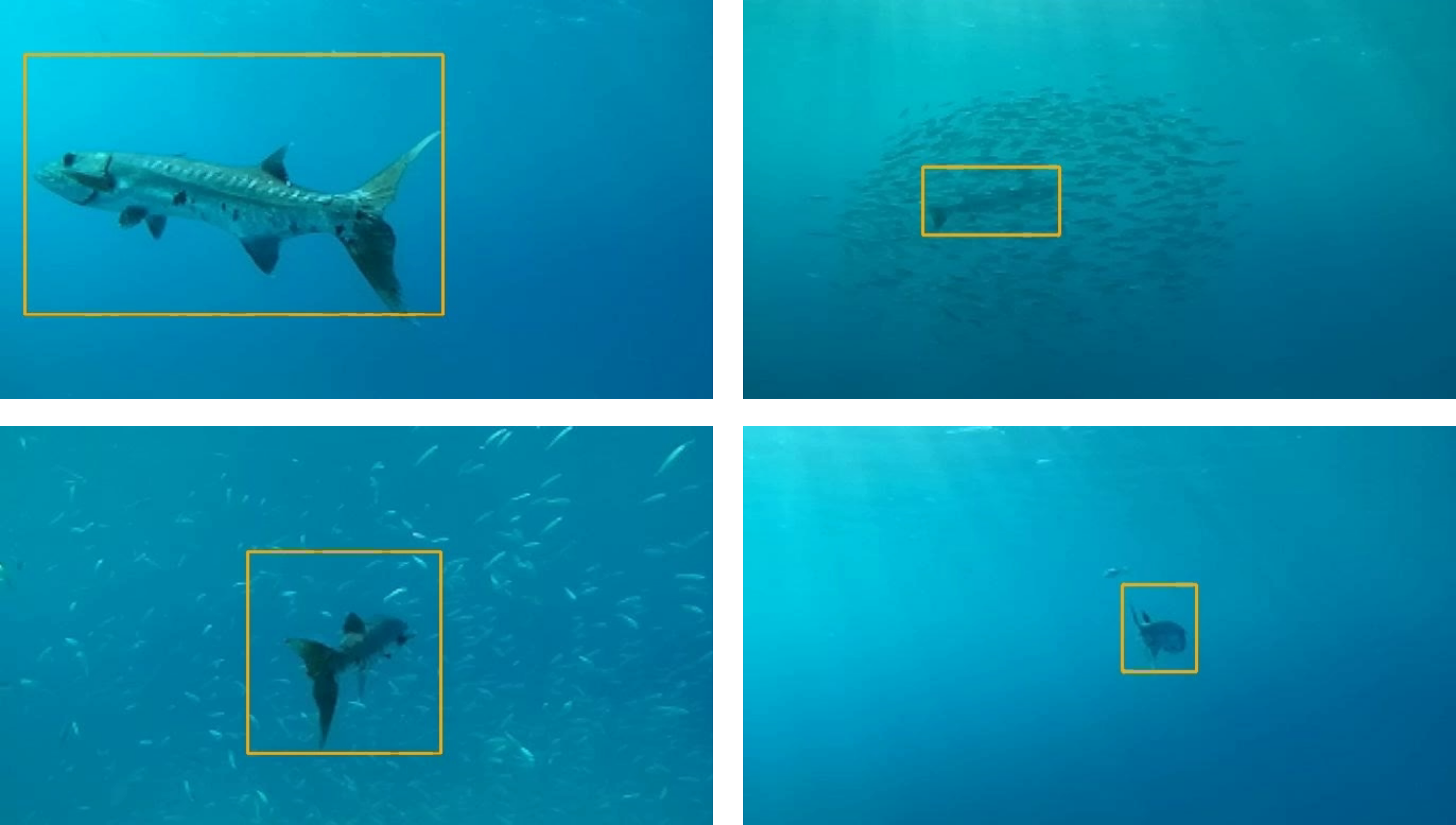}\\
    \vspace{0.2in}
    \includegraphics[width=.9\linewidth]{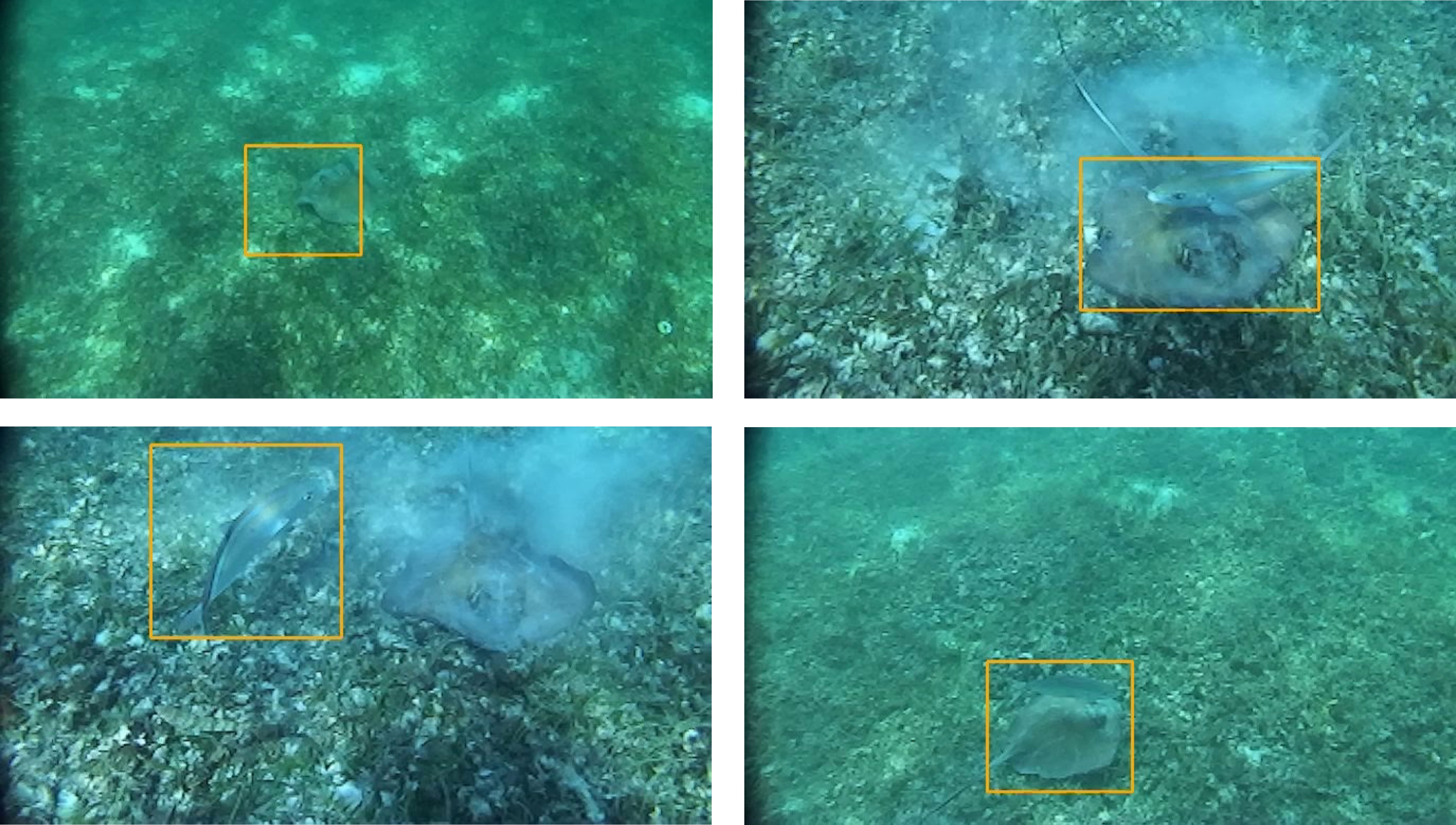}
    
    \caption{Still frames taken from autonomous tracking experiments. (TOP) CUREE follows a barracuda for five minutes, through significant changes in target appearance, and as the barracuda was partially occluded while swimming through schools of other fish. (BOTTOM) CUREE tracks a stingray for three minutes in a complex benthic scenario, though a symbiotic fish occasionally distracts the tracker.}
    \label{fig:barry_track}     \label{fig:ray_track}

\end{figure}
In this experiment we evaluated CUREE's ability to autonomously track moving objects in a reef environment using vision. Vision can be especially useful in shallow (<100m) reef environments since, during daytime, there is typically ample light and low turbidity, while sonar is generally ineffective due to severe multipathing in reef environments.  For the tracker we used a state-of-the-art semi-supervised tracker, SiamMask \cite{wang2019fast}, which we deployed on CUREE without requiring any additional training for the marine domain.  During the experiment, SiamMask was running at 15~fps on 360p images, fully on-board the vehicle.  CUREE's tracking controller operated in along 3 axes: yaw, surge, and heave to try to keep the bounding box produced by SiamMask centered within the image frame.  The heave and yaw actions are straightforward to compute from the bounding box position; the control input is proportional to the distance between the center of the bounding box and the center of the image.  However, since SiamMask is a monocular tracker, we cannot directly compute the surge feedback from the distance to the tracked target without performing registration to an object in a stereo depth image.  Instead, we opted for a simpler heuristic where the tracking controller attempts to keep the ratio of the width of the SiamMask bounding box to the width of the image constant.  In doing so, as the object grows larger in the image CUREE will move backwards away from it, and move closer if the object appears to get smaller.  

This tracker was deployed in two experiments near Tektite Reef in USVI.  In the first experiment, CUREE tracked a Barracuda as it swam in the water column above the reef, and in the second CUREE tracked a stingray swimming near the bottom of the reef.  In the Barracuda tracking experiment, CUREE was able to maintain its track for five minutes.  During the track, the barracuda swam through schools of small fish, as can be seen in Fig.~\ref{fig:barry_track} (upper-right, lower-left).  Another significant challenge for the tracker was the considerable change in appearance and apparent size of the barracuda depending on whether it was being viewed from the side (Fig.~\ref{fig:barry_track} top-left), or from behind (Fig.~\ref{fig:barry_track} bottom-right). This created an unforeseen emergent behavior within the CUREE tracking controller, where CUREE followed the barracuda much more closely while the fish was swimming directly away from CUREE.  However, despite being closely followed, the barracuda didn't seem perturbed by CUREE's presence, instead choosing to swim slowly through the water.

To further test the limits of the tracker, we selected a benthic target that had a more complex background. We were able to track a stingray, which can be seen in Fig.~\ref{fig:ray_track}, for several minutes. The stingray was followed by a symbiotic fish, which occasionally caused the tracker to follow the fish instead, though CUREE was able to eventually continue tracking the stingray when the fish returned to the stingray. This situation highlights the inherent challenges in this type of tracking. For further experiments and analysis refer to \cite{cai2023}.

In future work, the stereo-cameras on-board the AUV can be used to extract relative position, and hence an estimate of telemetry of the targets can be computed by merging those measurements with vehicle position estimates.

%%%%%%%%%%%%%%%%%%%%%%%%%%%%%%%%%%%%%%%%%%%%%%%%%%%%%%%%%%%%%%%%%%%%%%%%%%%%%%%%
\section{Hardware design and control}
\label{sec:hardware_design}
%%%%%%%%%%%%%%%%%%%%%%%%%%%%%%%%%%%%%%%%%%%%%%%%%%%%%%%%%%%%%%%%%%%%%%%%%%%%%%%%

\begin{figure}[!ht]
    \centering
    \includegraphics[width=.9\columnwidth]{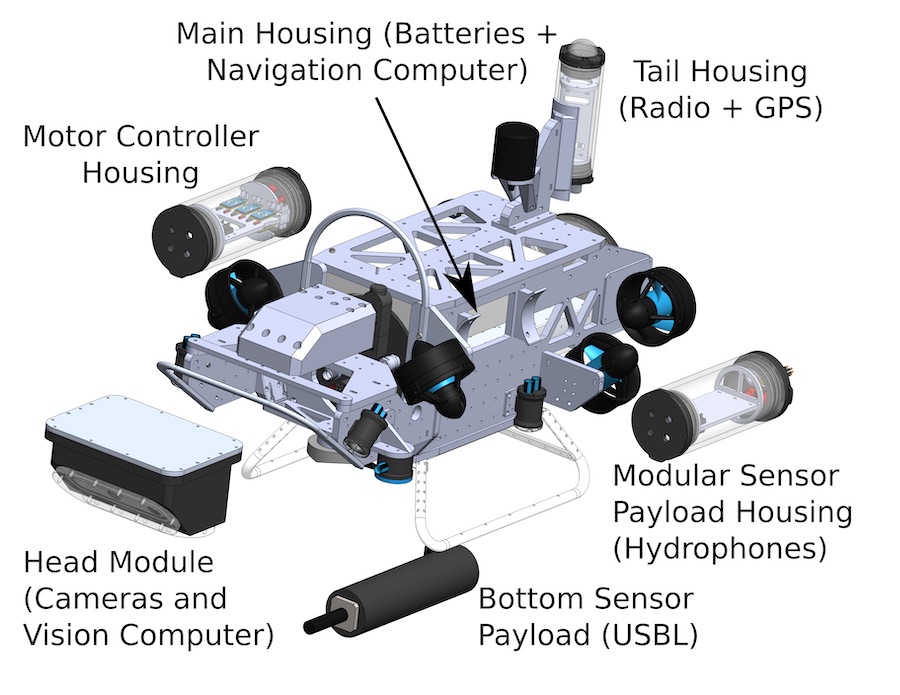}
  
    \caption{Exploded view of CUREE.  The head module contains the twin stereo pairs (survey cameras and tracking cameras) and can be swapped between a head with the tracking camera mounted forward facing and downward at $45^{\circ}$.  While the Sensor Payload housing currently holds the hydrophone recorder, it is connected to the main housing by ethernet and a power cable, and can therefore be easily swapped for other sensors (e.g. a water sample collector).}
    \label{fig:warpauv_exploded}
    \vspace{-0.4cm}
\end{figure}

CUREE is designed to be easily transportable and deployable by a single person anywhere in the world.  CUREE and its essential gear can fit in a pair of 27 $\times$ 16 $\times$ 9 inch Pelican cases, so it fits in the baggage allotment of most commercial airlines.  However, the dry weight of 55 lbs means that transporting CUREE this way will likely incur overweight baggage fees. CUREE can operate for multiple hours using one or two 14.8V 10Ah batteries from BlueRobotics, which can be carried on most commercial flights.

Aspects of CUREE's hardware design, specifically motor configuration and default camera configuration, were inspired by the Rangerbot \cite{zeldovich2018starfish}. However, CUREE is designed to be significantly more flexible, to support a larger range of scientific and engineering research, with modular sensor payloads and options for the head module that can be tailored to mission-specific camera configurations. CUREE utilizes a series of SubConn and BlueTrail connectors to enable this modularity.  Fig.~\ref{fig:warpauv_exploded} shows an expanded view of CUREE with its different housings and sensors. 

High maneuverability is necessary for low-altitude missions and collision avoidance in reef like environments and to track marine animals. Whereas stable forward motion is important for performing higher altitude survey tasks. To achieve both of these requirements, we use six Blue Robotics T200 thrusters in a configuration inspired by the Rangerbot \cite{zeldovich2018starfish}. Four thrusters are mounted in a quadrotor-like configuration but are then tilted 45-degrees inward to also allow for sideways motion. Two thrusters on the rear of the vehicle point directly aft, and control forward motion and yaw. This thruster configuration minimizes the amount of internal space taken up by the thrusters, keeps the vehicle streamlined in the forward direction, and provides full six degress of freedom. Polyurethane foam can be attached to attain neutral buoyancy, improve stability, and level trim in cases where surveying is more important than tracking.

CUREE's head module contains a downward-looking stereo camera, for substrate characterization or altitude estimation, and a forward-looking stereo camera suitable for animal tracking and collision avoidance, different head modules provide different angles, numbers, and types of cameras. In addition, the head module contains a NVIDIA Jetson Xavier AGX computer that runs all the neural networks and inference algorithms needed to implement the perception and path planning pipelines. CUREE also dissipates heat from the high-powered computers directly into the water by ensuring they are connected to the top of the aluminum head.

The main robot housing is a 6-inch cylindrical acrylic tube which contains the primary battery, networking equipment, and the primary control computer: a Raspberry Pi 4. A Waterlinked DVL and an IMU observe local changes in CUREE's position, orientation, altitude, and velocity, enabling dead reckoning positioning.  When deployed in combination with the USBL-equipped CuSA, CUREE also gets global positioning updates which are fused with dead reckoning based state estimates using an Extended Kalman Filter (EKF).  Altimetry or terrain relative navigation is obtained by fusing the vehicle's echosounders, DVLs, and stereo imaging based depth estimates. Both the control computer in the main housing and the Jetson in the robot's head run ROS, and all micro-controller code is written using Arduino, providing a fully open-source environment for development of science missions and implementation of new algorithms on CUREE.

CUREE's surface assistant (CuSA) has a GPS receiver and a USBL modem, allowing CUREE to obtain a global position reference while underwater.  CuSA also has a radio antenna, which allows high-bandwith information to be transmitted to scientists at the vehicles' base station.  However, all of the computation and power necessary to operate CUREE is onboard the vehicle, allowing it to operate independently from CuSA if desired.

%%%%%%%%%%%%%%%%%%%%%%%%%%%%%%%%%%%%%%%%%%%%%%%%%%%%%%%%%%%%%%%%%%%%%%%%%%%%%%%%
\section{Summary and Future Work}
\label{sec:conclusion}
%%%%%%%%%%%%%%%%%%%%%%%%%%%%%%%%%%%%%%%%%%%%%%%%%%%%%%%%%%%%%%%%%%%%%%%%%%%%%%%%

CUREE is a robot designed to explore underwater ecosystems, observe complex interactions between the organisms that live there and their habitats, and use these observations in real-time to adapt its behavior as an intelligent partner for marine science.  As a compact system designed to be deployed and operated by teams as small as a single person, CUREE can be taken anywhere in the world in checked luggage on commercial airlines and deployed without a need for significant supporting infrastructure.  In experiments in the U.S. Virgin Islands, we demonstrated how CUREE can be used to study coral reefs, by combining audio and visual observations of a coral reef to infer the preferred habitat of snapping shrimp, or by tracking a barracuda as it hunts above a reef.  As we continue the development of CUREE we plan to incorporate more active decision-making into its missions, enabling it to make informed decisions about which habitat types it should observe to gain the most information about the species that live there, or when to switch tracking targets.

\bibliographystyle{IEEEtran}
\bibliography{girdhar, smccammon, stewart}

\end{document}